\documentclass[11pt,a4paper]{article}
\usepackage[hyperref]{emnlp2020}
\usepackage{times}
\usepackage{todonotes}
\usepackage{latexsym}

\usepackage{microtype}

\usepackage{amsmath}
\usepackage{amssymb}
\usepackage{hyperref}
\usepackage{graphicx}
\usepackage{ifthen}
\usepackage[ruled]{algorithm2e}
\usepackage{subcaption}
\usepackage{tabulary}
\usepackage{svg}
\usepackage{booktabs} %

\definecolor{darkgreen}{HTML}{024b30}
\definecolor{darkgray}{HTML}{474747}

\usepackage{url}

\aclfinalcopy 
\def\aclpaperid{1490} 

\usepackage{cleveref}

\crefformat{section}{\S#2#1#3} %
\crefformat{subsection}{\S#2#1#3}
\crefformat{subsubsection}{\S#2#1#3}

\newcommand{\autoencoder}{\operatorname{\mathcal{A}}}

\newcommand{\discretespace}{\mathcal{X}}

\newcommand{\encoder}{\mathrm{enc}}
\newcommand{\decoder}{\mathrm{dec}}
\newcommand{\loss}{\mathcal{L}}
\newcommand{\embtransformation}{\Phi}
\newcommand{\discriminator}{\mathrm{disc}}

\newcommand{\lambdaadv}{\lambda_{\mathit{adv}}}
\newcommand{\lambdaclf}{\lambda_{\mathit{sty}}}

\usepackage{accents}

\newcommand{\etoescratch}{S2S-Scratch\xspace}
\newcommand{\etoept}{S2S-Pretrain\xspace}
\newcommand{\etoemlp}{S2S-MLP\xspace}
\newcommand{\etoefreeze}{S2S-Freeze\xspace}

\title{Plug and Play Autoencoders for Conditional Text Generation}

\author{
Florian Mai$^{\dagger}$$^{\spadesuit}$ \quad
Nikolaos Pappas$^{\clubsuit}$ \quad
Ivan Montero$^{\clubsuit}$ \quad \\
\textbf{Noah A. Smith}$^{\clubsuit}$$^{\diamondsuit}$ \quad 
\textbf{James Henderson}$^{\dagger}$
\vspace{3mm}
\\
  $^{\dagger}$Idiap Research Institute, Martigny, Switzerland  \\
  $^{\spadesuit}$EPFL, Lausanne, Switzerland \\
  $^{\clubsuit}$University of Washington, Seattle,  WA, USA \\
   $^{\diamondsuit}$Allen Institute for Artificial Intelligence, Seattle, WA, USA\\
    {\tt \{florian.mai,james.henderson@idiap.ch\} }\\
   {\tt \{npappas,ivamon,nasmith@cs.washington.edu\} } 
}
\date{}

\begin{document}
\maketitle
 
\begin{abstract}
  Text autoencoders are commonly used for conditional generation tasks such as style transfer.   We propose methods which are plug and play, where any pretrained autoencoder can be used, and only require learning a mapping within the autoencoder's embedding space, training embedding-to-embedding (\emph{Emb2Emb}).  This reduces the need for labeled training data for the task and makes the training procedure more efficient.  Crucial to the success of this method is a loss term for keeping the mapped embedding on the manifold of the autoencoder and a mapping which is trained to navigate the manifold by learning offset vectors.  Evaluations on style transfer tasks both with and without sequence-to-sequence supervision show that our method performs better than or comparable to strong baselines while being up to four times faster.

\end{abstract}

\section{Introduction}

Conditional text generation\footnote{We use this term to refer to text generation conditioned on \emph{textual} input.} %
encompasses a large number of natural language processing tasks such as text simplification  \cite{nisioi-etal-2017-exploring,zhang2017sentence}, summarization \cite{rush15,nallapati-etal-2016}, machine translation  \cite{bahdanau2014neural,kumar2018mises} and style transfer  \cite{shen2017style,fu2018style}. When training data is available, the state of the art includes encoder-decoder models with an attention mechanism  \citep{bahdanau2014neural,vaswani2017attention} which are both extensions of the original sequence-to-sequence framework with a fixed bottleneck introduced by \citet{sutskever2014sequence}. Despite their success, these models are costly to train and require a large amount of parallel data. 
\begin{figure}[ht]
\centering
\hspace{-2mm}\includegraphics[width=0.42\textwidth]{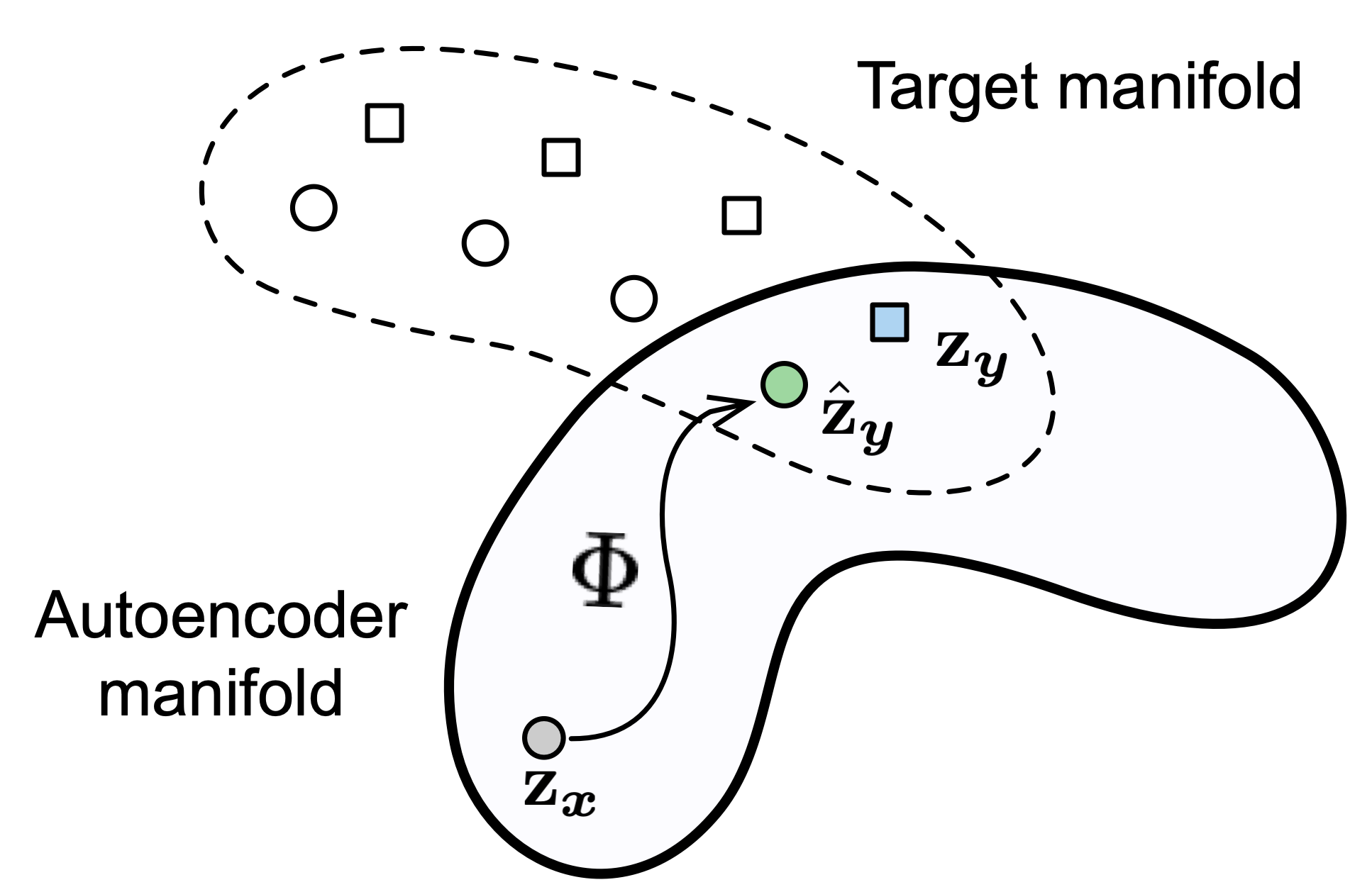}
\vspace{-1mm}
\caption{The \textit{manifold} of a text autoencoder is the %
low-dimensional region of the high-dimensional embedding space where texts are actually embedded. 
 The example shows the %
 mapping of a source sequence $\boldsymbol{x}$ with embedding $\mathbf{z}_{\boldsymbol{x}}$ %
 to $\bf{z}_{\boldsymbol{y}}$, which is the embedding of  target sequence $\boldsymbol{y}$  such that it reflects the target  manifold.
\label{fig:overview}}
\vspace{-3mm}
\end{figure}
\begin{figure*}
\centering
\includegraphics[width=0.97\textwidth]{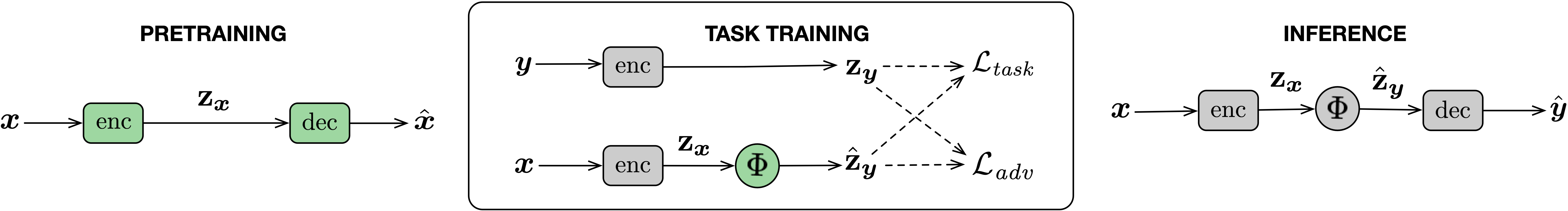}
\caption{%
High-level view of the supervised variant of our framework \emph{Emb2Emb}.  \textit{Left}:  we pretrain an autoencoder on (unannotated) text, which transforms an input sentence $\boldsymbol{x}$ into an embedding $\bf{z_{\boldsymbol{x}}}$ and uses it to predict a reconstruction $\boldsymbol{\hat{x}}$ of the input sentence.  \textit{Center}:  using the (frozen, hence depicted in \textcolor{darkgray}{gray}) autoencoder, we learn a mapping $\Phi$ (trained, hence depicted in \textcolor{darkgreen}{green}) from the autoencoder's embedding of an input $\bf{z_{\boldsymbol{x}}}$ to the embedding $\bf{z_{\boldsymbol{y}}}$ of the output sentence $\boldsymbol{y}$. The training objective consists of two losses: $\loss_{\mathit{task}}$ enforces the predicted output embedding to be close to the true output embedding, and $\loss_{\mathit{adv}}$ is an adversarial loss term that enforces the output embedding to be on the manifold of the autoencoder. \textit{Right}:  at inference time, $\Phi$ is composed between the autoencoder's encoder and decoder to transform input sentence $\boldsymbol{x}$ to output sentence  $\boldsymbol{\hat{y}}$.  Not shown:  the unsupervised variant where only $\boldsymbol{x}$ (not $\boldsymbol{y}$) sequences are available in task training (Section \ref{eq:unsupervised-loss}).
\label{fig:framework_stages}}
\end{figure*} 

Yet parallel data is scarce for conditional text generation problems, necessitating unsupervised solutions. 
Text autoencoders~\cite{bowman2015generating} have proven useful for a particular subclass of unsupervised problems that can be broadly defined as style transfer, i.e., changing the style of a text in such a way that the content of the input is preserved. Examples include sentiment transfer~\cite{shen2017style}, sentence compression~\cite{fevry2018unsupervised}, and neural machine translation~\cite{artetxe2017unsupervised}. 
Most existing methods specialize autoencoders to the task by conditioning the decoder on the style attribute of interest~\cite{subramanian2018multiple,logeswaran2018content}, assuming the presence of labels during training of the autoencoder.
The main drawback of this approach is that
it cannot leverage pretraining on \textbf{unlabeled data}, which is probably the most important factor for widespread progress in supervised NLP models in recent years in text analysis~\cite{Peters:2018, radford2019language, devlin-etal-2019-bert}
and  generation tasks~\cite{pmlr-v97-song19d,varis-bojar-2019-unsupervised}. There are no style transfer methods, to the best of our knowledge, that were designed to leverage autoencoder pretraining, and only few can be used in this way~\cite{shen2019latent, wang2019controllable}. 

In this paper, we propose an autoencoder-based framework that
is \emph{plug and play},\footnote{This term is borrowed from studies %
on unconditional text generation with a specific attribute \cite{duan2019pre}.}  %
meaning it can be used with any pretrained autoencoder, and thus can benefit from pretraining.
Instead of learning conditional text generation in the discrete, high-dimensional space where texts are actually located, our method, called \textbf{Emb2Emb}, 
does all learning in the low-dimensional continuous embedding space, on the \emph{manifold} of a pretrained text autoencoder (see Figure~\ref{fig:overview}).  The result of learning is simply a mapping from input embedding to output embedding.
Two crucial model choices enable effective learning of this mapping.
First, an adversarial loss term encourages the output of the mapping to remain on the manifold of the autoencoder, to ensure effective generation with its decoder.  Second, our neural mapping architecture is designed to learn offset vectors that are added to the input embedding, enabling the model to make small adjustments to the input to solve the task. %
Lastly, we propose two conditional generation models %
based on our framework, one for supervised style transfer (Section \ref{sec:supervised}) and the other for unsupervised style transfer (Section \ref{sec:unsupervised}) 
that implement the criteria of content preservation and attribute transfer directly on the autoencoder manifold.

We evaluate on two style transfer tasks for English.  On text simplification (Section \ref{sec:text_simplification}), where supervision is available, we find that our approach outperforms conventional end-to-end training of models with a fixed-size ``bottleneck'' embedding  (like an autoencoder) while being about four times faster. %
On unsupervised sentiment transfer (Section \ref{sec:style_transfer}),
where no parallel sentence pairs are available to supervise learning, and where models with a fixed-size bottleneck are a common choice,   
Emb2Emb preserves the content of the input sentence better than a state-of-the-art method while achieving comparable transfer performance.
Experimentally, we find that our method, due to being plug and play, achieves performances close to the full model when only 10\% of the labeled examples are used, demonstrating the importance of pretraining for this task.

Our contributions can be summarized as follows:
\begin{itemize}
    \item Our proposed framework Emb2Emb reduces conditional text generation tasks to mapping between continuous vectors in an autoencoder's embedding space.
    \item We propose a neural architecture and an adversarial loss term that facilitate learning this mapping.%
    \item We evaluate two new conditional generation models for generation tasks with and without parallel examples as supervision.%
    \item We demonstrate that our model benefits substantially from pretraining on large amounts of unlabeled data, reducing the need for large labeled corpora. %
\end{itemize}

\section{Proposed Framework} \label{sec:framework}

The key idea of our framework is to reduce discrete sequence-to-sequence tasks to a continuous embedding-to-embedding regression problem.
Our Emb2Emb framework for conditional generation based on pretrained autoencoders (Figure \ref{fig:framework_stages}) encompasses learning sequence-to-sequence tasks both where parallel input/output sentence pairs are available (``supervised'') and where they are not (``unsupervised'').
Given a pretrained autoencoder (left of Figure~\ref{fig:framework_stages}, Section~\ref{sec:autoencoders})
we use its encoder to encode both input and, during supervised training, the output sequence (Section \ref{sec:supervised}).\footnote{
Our code is available at: \url{https://github.com/florianmai/emb2emb}}
We then learn a continuous mapping $\Phi$ from input sequence embeddings to output sequence embeddings (center of Figure~\ref{fig:framework_stages}).
In the unsupervised case (Section~\ref{sec:unsupervised}), the task loss reflects the objectives of the task, in our experiments consisting of two terms, one to encourage content preservation and the other to encourage style transfer. In both supervised and unsupervised cases, the task-specific loss $\loss_{\mathit{task}}$ is combined with an adversarial term $\loss_{\mathit{adv}}$ that encourages the output vector to stay on the autoencoder's manifold (see Figure~\ref{fig:overview}; Section~\ref{sec:adv_loss}), so that the complete loss function is:
\begin{equation}
  \label{eq:loss}
\loss = \loss_{\mathit{task}} + \lambdaadv \loss_{\mathit{adv}}.
\end{equation}
At inference time (right of Figure~\ref{fig:framework_stages}; Section~\ref{sec:inference}), the decoder from the pretrained autoencoder's decoding function is composed with $\Phi$ and the encoder to generate a discrete output sentence $\boldsymbol{\hat{y}}$ conditioned on an input sentence $\boldsymbol{x}$:
\begin{equation}
    \boldsymbol{\hat{y}} = (\decoder \circ \embtransformation \circ \encoder)(\boldsymbol{x})
\end{equation}

\subsection{Text Autoencoders} \label{sec:autoencoders}

The starting point for our approach is an autoencoder $\autoencoder =  \decoder \circ \encoder$  %
trained to map an input sentence to itself, i.e., $\autoencoder(\boldsymbol{x}) = \boldsymbol{x}$.
Letting $\discretespace$ denote the (discrete) space of text sequences, the encoder $\encoder : \discretespace \rightarrow \mathbb{R}^d$ produces an intermediate 
continuous vector representation (embedding), which is turned back into a sequence by the decoder $\decoder : \mathbb{R}^d \rightarrow \discretespace$.  
Note that an autoencoder can, in principle, be pretrained on a very large dataset, because it does not require any task-related supervision. 

While our framework is compatible with any type of autoencoder, in practice, learning the mapping $\embtransformation$ will be easier if the embedding space is smooth. How to train smooth text autoencoders is subject to ongoing research \citep{bowman2015generating, shen2019latent}.
In this work, we will focus on denoising recurrent neural network autoencoders (\citealp{vincent2010stacked, shen2019latent}; see Appendix \ref{app:experimental-details}). However, any advancement in this research direction will directly benefit our framework.

\begin{figure}[ht]
\vspace{-2mm}
\centering
 \hspace{-2mm}\includegraphics[width=0.85\columnwidth]{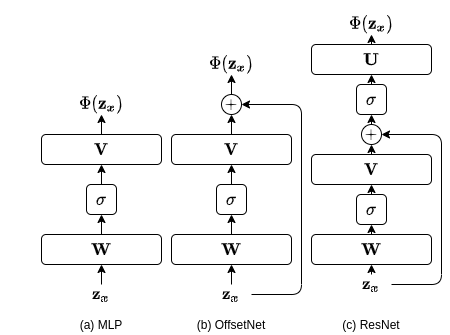}
 \vspace{-2mm}
  \caption{Illustration of the three neural architectures (1 layer) considered in this study. OffsetNet (b) differs from ResNet (c) in that there is no non-linear activation after the skip-connection (+), satisfying the notion of computing an offset vector that is added to the input. %
  } \vspace{-3mm}
  \label{fig:neural-architectures}
\end{figure}
\subsection{Mapping Function $\Phi$}\label{sec:navigating-the-manifold}

A common choice for the mapping $\embtransformation$ to learn a regression task would be a $k$-layer MLP \cite{rumelhard86},
which transforms the input $\mathbf{z}_{\boldsymbol{x}} \in \mathbb{R}^{d}$ as:
\begin{align} 
\mathbf{y}^{(0)} & = \mathbf{z}_{\boldsymbol{x}} \\
\forall j \in \{1, \ldots, k\}, \quad \mathbf{y}^{(j)} & = \sigma(\mathbf{W}^{(j)} \mathbf{y}^{(j-1)}) \label{eq:recur} \\
\embtransformation(\mathbf{z}_{\boldsymbol{x}}) & = \mathbf{W}^{(k)}\mathbf{y}^{(k)},
\end{align}
where $\mathbf{W}^{(j)} \in \mathbb{R}^{d \times d}$ are linear transformations and $\sigma$ denotes a non-linear activation function.
The linear transformation at the output layer allows $\embtransformation$ to match the unbounded range of the regression task. Note that we have suppressed the bias terms of the transformations for clarity. %
In past work \citep{shen2019latent}, a mapping function was chosen with a specific form,
\begin{equation}
    \phi(\mathbf{z}) = \mathbf{z} - \mathbf{v}_{1} + \mathbf{v}_2,
\end{equation}
where the ``offset'' vectors $\mathbf{v}_1$ and $\mathbf{v}_2$ correspond to encodings of the input style and the output style, computed as the average of sentences with the respective style. Because dimensions not relating to style information cancel each other out, the output remains close to the input. However,  this model lacks generality because the offset vectors are independent of the input. %

We propose a mapping which incorporates the notion of an offset vector, but is conditioned on the input.
Each layer of the $\embtransformation$ network moves through the embedding space by
an input-specific offset, computed using a ``skip'' connection at each layer:
\begin{equation}
  \mathbf{y}^{(j)} = \mathbf{y}^{(j-1)} + \underbrace{\mathbf{V}^{(j)} \sigma(\mathbf{W}^{(j)}\mathbf{y}^{(j-1)})}_{\mbox{offset $j$}}  \label{eq:offsetnet}.
\end{equation}
$\mathbf{V}^{(j)}, \mathbf{W}^{(j)} \in \mathbb{R}^{d \times d}$ again denote linear transformations.
Unlike the MLP, the skip connections 
bias  the output to be close to the input.
We refer to this architecture as \textbf{OffsetNet}.

Note that a residual network \citep{he2016deep} corresponds to Equation~\ref{eq:offsetnet} but with an additional non-linear transformation (typically of bounded range) applied to the output of Equation~\ref{eq:offsetnet}, again necessitating a linear transformation at the output layer. %
However, transforming the output all together would defeat the purpose of navigating the manifold with learned offsets.
Figure~\ref{fig:neural-architectures} illustrates the differences.   %
Our experiments (Section~\ref{sec:exp:wiki:analyis}) validate the design choice for OffsetNet.

\subsection{Supervised Task Loss $\loss_{\mathit{task}}$} 
\label{sec:supervised}

For supervised tasks, a collection of parallel sentence pairs $\left\langle (\boldsymbol{x}_i, \boldsymbol{y}_i)\right \rangle_{i=1}^N$  is available as supervision for the conditional generation task.  After pretraining the autoencoder, $\encoder$ is available to transform the training data into pairs of vectors $(\mathbf{z}_{\boldsymbol{x}_i} {=} \encoder(\boldsymbol{x}_i), \mathbf{z}_{\boldsymbol{y}_i} {=} \encoder(\boldsymbol{y}_i))$, giving us: %
\begin{equation}
  \loss_{\mathit{task}} =
  \frac{1}{N} \sum_{i=1}^N \loss_{\mathit{emb}}(\embtransformation(\mathbf{z}_{\boldsymbol{x}_i}; \boldsymbol{\theta}), \mathbf{z}_{\boldsymbol{y}_i}). %
    \label{eq:suptrain}
\end{equation}

The multivariate regression in Equation~\ref{eq:suptrain} requires that we specify a loss function, $\loss_{\mathit{emb}}$, which should reflect semantic relatedness in the autoencoder's embedding space.
For sentence and word embeddings, past work has concluded that cosine distance is preferable to Euclidean distance (i.e., mean squared error) %
in such settings~\cite{xing2015normalized, bhat2019margin},  which agreed with our preliminary experiments; hence, we adopt cosine distance for  the task-specific loss $\loss_{\mathit{emb}}$. %

\citet{kumar2018mises} showed that another alternative, the Von Mises-Fisher loss, was preferable in learning to generate continuous word vector outputs.  Their loss is not applicable in our setting, because the embedding space of an autoencoder is not unit-normalized like word vectors typically are.  Therefore, we employ cosine loss and leave the exploration of other regression losses to future work.

\subsection{Unsupervised Task Loss $\loss_{\mathit{task}}$} \label{sec:unsupervised}

In the unsupervised case, we do not have access to parallel sentence pairs $(\boldsymbol{x}, \boldsymbol{y})$. Instead, we have a collection of sentences labeled with their
style attribute (e.g., sentiment), here denoted $\left\langle(\boldsymbol{x}_i, a_i)\right\rangle_{i{=}1}^N$. %
The goal of the task is twofold \citep{logeswaran2018content}:  preserve the content of the input and match the desired value for the style attribute.  We view this as a tradeoff, defining $\loss_{\mathit{task}}$ as an interpolation between loss terms for each.
With $\hat{\mathbf{z}}_{\boldsymbol{x}_i} {=} \embtransformation(\mathbf{z}_{\boldsymbol{x}_i}; \boldsymbol{\theta})$,
for the unsupervised case we have:
\begin{align}
  \loss_{\mathit{task}} = 
\hspace{-1mm}\lambdaclf \loss_{\mathit{sty}}(\hat{\mathbf{z}}_{\boldsymbol{x}_i}) 
  + (1-\lambdaclf) \loss_{\mathit{cont}}(\hat{\mathbf{z}}_{\boldsymbol{x}_i}, \mathbf{z}_{\boldsymbol{x}_i}). \label{eq:unsupervised-loss}
\end{align}
where $\loss_{\mathit{cont}}$ and $\loss_{\mathit{sty}}$ are described in the following. Lastly, we describe an inference-time method that can improve the loss after applying the mapping even further. 

\paragraph{Content preservation.}
We encourage the output to stay close to the input, on the assumption that embeddings are primarily about semantic content.
To this end, we choose
$\loss_{\mathit{cont}}(\embtransformation(\mathbf{z}_{\boldsymbol{x}_i}; \boldsymbol{\theta}), \mathbf{z}_{\boldsymbol{x}_i})$ to be cosine distance.

\paragraph{Style.}
Following previous approaches~\cite{engel2017latent, liu2019revision, wang2019controllable}, our style objective requires that we pretrain a (probabilistic) classifier that predicts the style attribute value from the (fixed) autoencoder's embedding.  The classifier is then frozen (like the autoencoder) and our minimizing objective requires the output of our method to be classified as the target style. %
Formally, in a preliminary step, we train a style classifier $\mathbf{c} : \mathbb{R}^d \rightarrow \{0,1\}$ on the embeddings of the autoencoder to predict one of the attributes (labeled as 0 and 1, respectively). We then freeze the classifier's parameters, and encourage $\embtransformation$ to produce outputs of the target attribute ($y {=} 1$) via a negative log-likelihood loss:
\begin{align*}
    \loss_{\mathit{sty}}(\embtransformation(\mathbf{z}_{\boldsymbol{x}_i}; \boldsymbol{\theta}), \mathbf{z}_{\boldsymbol{x}_i}) & = - \log ( \mathbf{c}(\embtransformation(\mathbf{z}_{\boldsymbol{x}_i}; \boldsymbol{\theta}))).
\end{align*}

\paragraph{Inference Time.} \label{sec:inference}
The mapping $\Phi$ is trained to try to optimize the objective \eqref{eq:loss}.
In the unsupervised %
case, we can actually verify at test time whether it has succeeded, since nothing is known at training time that is not also known at test time (i.e.,  no labeled output).
We propose a second stage of the mapping for the unsupervised case which corrects any suboptimality.
We apply fast gradient iterative modification (FGIM;~\citealp{wang2019controllable}) to improve the predicted embeddings further.
Formally, we modify the predicted embedding $\hat{\bf{z}}_{\boldsymbol{x}_i} {=} \embtransformation(\mathbf{z}_{\boldsymbol{x}_i}; \boldsymbol{\theta})$ as
\begin{align*}
    \hat{\hat{\bf{z}}}_{\boldsymbol{x}_i} = \hat{\bf{z}}_{\boldsymbol{x}_i} + \omega \nabla_{\hat{\bf{z}}_{\boldsymbol{x}_i}} \loss(\hat{\bf{z}}_{\boldsymbol{x}_i}, \bf{z_{\boldsymbol{x}_i}}),
\end{align*} 
where $\omega$ is the stepsize hyperparameter. This step is repeated for a fixed number of steps or until $\mathbf{c}(\hat{\hat{\bf{z}}}_{\boldsymbol{x}_i}) > t$, where, $\mathbf{c}$ is the style classifier from above, and $t \in [0,1]$ denotes some threshold.

\citet{wang2019controllable} use this method to modify the input embedding $\mathbf{z_{\boldsymbol{x}_i}}$ by only following the gradient of the classifier $\mathbf{c}$,
i.e., $\hat{\hat{\bf{z}}}_{\boldsymbol{x}_i} = \bf{z}_{\boldsymbol{x}_i} + \omega \nabla_{\bf{z}_{\boldsymbol{x}_i}} - \log \bf{c}(\mathbf{z}_{\boldsymbol{x}_i})$. In contrast, our variant takes the entire loss term into account, including the adversarial term that encourages embeddings that lie on the manifold of the autoencoder, which we explain next.

\subsection{Adversarial Loss $\loss_{\mathit{adv}}$}\label{sec:adv_loss}

Recall that at test time the output of $\embtransformation$ is the input to the pretrained decoding function $\decoder$.  Even for supervised training, we do not expect to obtain zero loss during training (or to generalize perfectly out of sample), so there is a %
concern that the output of $\embtransformation$ will be quite different from the vectors $\decoder$ was trained on (during pretraining).  In other words, there is no guarantee that $\embtransformation$ will map onto the manifold of the autoencoder.

To address this issue, we propose an adversarial objective that encourages the output of  $\embtransformation$ to remain on the manifold. 
Our method is similar to the ``realism'' constraint of \citet{engel2017latent}, who train a discriminator to distinguish between latent codes drawn from a prior distribution (e.g., a multivariate Gaussian) and the latent codes actually produced by the encoder. Instead of discriminating against a prior (whose existence we do not assume), we discriminate against the embeddings produced by $\embtransformation$.
We build on the adversarial learning framework of \citet{goodfellow2014generative} to encourage the transformation $\embtransformation$ to generate output embeddings indistinguishable from the embeddings produced by the encoder $\encoder$. 

Formally, let $\discriminator$ be a (probabilistic)  binary classifier responsible for deciding whether a given embedding was generated by $\encoder$  or $\embtransformation$. 
The discriminator is trained to distinguish between embeddings produced by $\encoder$ and embeddings produced by $\embtransformation$:
\begin{align}
\label{eq:discriminator}
    \max_{\discriminator} \sum_{i=1}^N &\log (\discriminator(\mathbf{z}_{\boldsymbol{\tilde{y}}_i}))  + \log ( \overline{\discriminator}(\embtransformation(\mathbf{z}_{\boldsymbol{x}_i}))
\end{align}
\noindent where $\discriminator(\mathbf{z})$ denotes the probability of vector $\mathbf{z}$ being produced by $\encoder$ and $\overline{\discriminator}(\mathbf{z}) = 1{-}\discriminator(\mathbf{z})$.
The mapping $\embtransformation$ is trained to ``fool'' the discriminator:
\begin{equation} 
\loss_{\mathit{adv}}(\embtransformation(\mathbf{z}_{\boldsymbol{x}_i}); \boldsymbol{\theta}) = -\log (\discriminator(\embtransformation(\mathbf{z}_{\boldsymbol{x}_i}); \mathbf{\theta})) 
\label{eq:adversarial}
\end{equation}

Training the discriminator requires encoding negatively sampled sentences, $\mathbf{z}_{\boldsymbol{\tilde{y}}_i} {=} \encoder(\tilde{y}_i)$, where we want these sentences to contrast with the output of the mapping $\embtransformation(\mathbf{z}_{\boldsymbol{x}_i})$.  For the supervised case, we achieve this by taking the negative samples from the target sentences of the training data, $\tilde{y}_i {=} y_i$.
In the unsupervised case, $\tilde{y}_i$ are sampled randomly from the data.

The mapping $\embtransformation$ is trained according to the objective in \eqref{eq:loss}, in which $\loss_{\mathit{adv}}$ depends on training the discriminator $\discriminator$ according to \eqref{eq:discriminator}.
In practice, we alternate between batch updates to $\embtransformation$ and $\discriminator$. %
Our experiments in Section~\ref{sec:experiments} will explore sensitivity to $\lambdaadv$, finding that it has a large effect.  In practical applications, it should therefore be treated as a hyperparameter.  %

\subsection{Summary}\label{sec:summary}

Our framework is \emph{plug and play}, since it is usable with any pretrained autoencoder. Unlike previous methods by \citet{shen2019latent} and  \citet{wang2019controllable}, which are specific to style transfer and do not learn a function (like $\embtransformation$ in Emb2Emb), ours can, in principle, be used to learn a mapping from any sort of input data to text, as long as the desired attributes of the generated text can be expressed as a loss function that is tied to the autoencoder manifold. In this study, we apply it to supervised and unsupervised text style transfer.
The key component is the mapping function $\embtransformation$, which is trained via a regression loss (plus auxiliary losses) to map from the embedding of the input sequence to the embedding of the output sequence. Learning the function is facilitated through the proposed \textbf{OffsetNet} and an adversarial loss term that forces the outputs of the mapping to stay on the manifold of the autoencoder.

\section{Experiments}\label{sec:experiments}

We conduct controlled experiments to measure the benefits of the various aspects of our approach.  First, we consider a supervised sentence simplification task and compare our approach to others that use a fixed-size representational bottleneck, considering also the model's sensitivity to the strength of the adversarial loss and the use of OffsetNet (Section~\ref{sec:text_simplification}).  We then turn to an unsupervised sentiment transfer task, first comparing our approach to other methods that can be considered ``plug and play''  and then 
investigating
the effect of plug and play when only a little labeled data is available (Section~\ref{sec:style_transfer}).
We note that the current state of the art is based on transformers \cite{vaswani2017attention}; since our aim is to develop a general-purpose framework and our computational budget is limited, we focus on controlled testing of components rather than achieving state-of-the-art performance.

\paragraph{Autoencoder.}
In all our experiments, we use a one-layer LSTM as encoder and decoder, respectively. We pretrain it on the text data of the target task as a denoising autoencoder (DAE; \citealp{vincent2010stacked}) with the noise function from \citet{shen2019latent}.
Additional training and model details can be found in Appendix~\ref{app:experimental-details}.

\subsection{Sentence Simplification}\label{sec:text_simplification}

Sentence simplification provides a useful testbed for the supervised variant of our approach.  The training data contains pairs of  input and output sentences $(\boldsymbol{x}_i,\boldsymbol{y}_i)$, where $\boldsymbol{x}_i$ denotes the input sentence in English and $\boldsymbol{y}_i$ denotes the output sentence in simple English. %
We evaluate on the English WikiLarge corpus introduced by \citet{zhang2017sentence}, which consists of 296,402 training pairs, and development and test datasets adopted from \citet{xu2016optimizing}.
Following convention, we report two scores:
BLEU~\cite{papineni2002bleu}, which correlates with grammaticality \citep{xu2016optimizing}, and SARI~\cite{xu2016optimizing}, found to correlate well with human judgements of simplicity.  We also compare training runtimes.

\subsubsection{Comparison to Sequence-to-Sequence} \label{sec:exp:wiki:end2end}

Our first comparisons focus on models that, like ours, use a fixed-size encoding of the input. %
Keeping the autoencoder architecture fixed (i.e., the same as our model), we consider variants of the sequence-to-sequence model of \citet{sutskever2014sequence}.\footnote{On this task, much stronger performance than any we report has been achieved using models without this constraint \citep{mathews2018simplifying,zhang2017sentence}.  Our aim is not to demonstrate superiority to those methods; the fixed-size encoding constraint is of general interest because (i) it is assumed in other tasks such as unsupervised style transfer and (ii) it is computationally cheaper.} All of these models are trained ``end-to-end,'' minimizing token-level cross-entropy loss.  The variants are:
\begin{table}[htp]
    \centering
  \def\arraystretch{1.1}\tabcolsep=10pt    
    \begin{tabular}{l|r|r|r}
        \toprule
        \textbf{Model} & \textbf{BLEU} & \textbf{SARI} & \textbf{Time} \\
        \toprule
        \hline
        \etoescratch & 3.6 & 15.6 & 3.7$\times$ \\
        \etoept & 5.4 & 16.2 & 3.7$\times$ \\
        \etoemlp & 10.5 & 17.7 & 3.7$\times$ \\
        \etoefreeze & 23.3 & 22.4 & 2.2$\times$ \\
        \hline
        Emb2Emb & \textbf{34.7} & \textbf{25.4} & 1.0$\times$ \\ \bottomrule
    \end{tabular}
    \caption{Text simplification performance of model variants of end2end training on the test set.
    ``Time'' is wall time of one training epoch, relative to our model, Emb2Emb. %
    }
    \label{tab:end2end}
    \vspace{-3mm}
\end{table}
\begin{itemize}
    \item \textbf{\etoescratch}: trains the model from scratch.
    \item \textbf{\etoept}: uses a pretrained DAE and finetunes it.
    \item \textbf{\etoemlp}: further adds the trainable mapping $\Phi$ used in our approach.
    \item \textbf{\etoefreeze}: freezes the pretrained autoencoder parameters, which we expect may help with the vanishing gradient problem arising in the rather deep \etoemlp variant.
\end{itemize}

\noindent For all the models, we tuned the learning rate hyperparameter in a comparable way and trained with the ADAM optimizer by \citet{kingma2014adam} (more details in the Appendix \ref{sec:sim_details}).

\paragraph{Results.}
Table~\ref{tab:end2end} shows test-set performance and the runtime of one training epoch relative to our model (Emb2Emb).  
First, note that the end-to-end models are considerably more time-consuming to train. \etoefreeze is not only more than two times slower per epoch than Emb2Emb, but we find it to also require 14 epochs to converge (in terms of validation performance), compared to 9 for our model. Turning to accuracy, as expected, adding pretraining and the MLP to \etoescratch does improve its performance, but freezing the autoencoder (\etoefreeze) has an outsized benefit.  This observation may seem counter to the widely seen success of finetuning across other NLP scenarios, in particular with pretrained transformer models like BERT~\citep{devlin-etal-2019-bert}. However, finetuning does not always lead to better performance. For instance, \citet{peters-etal-2019-tune} not only find the LSTM-based ELMo~\citep{Peters:2018} difficult to configure for finetuning in the first place, but also observe performances that are often far lower than when just freezing the parameters. Hence, our results are not entirely unexpected. 
To further eliminate the possibility that the finetuned model underperformed merely because of improper training, we verified that the training loss of \etoept is indeed lower than that of \etoefreeze. 
Moreover, the poor performance is also unlikely to be a problem of overfitting, because we mitigate this via early stopping.
This suggests that the differences are largely due to the generalization abilities coming from the pretraining, which is partly forgotten when finetuning the entire model on the target task.
Our results thus support the hypothesis of \citet{mathews2018simplifying} that fixed-size bottlenecks and deeper networks make end-to-end learning harder. In contrast, training Emb2Emb even outperforms the best end-to-end model, \etoefreeze.

\begin{figure}[t]
\vspace{-7mm}
\centering
\hspace{-2mm}\includegraphics[width=1.05\columnwidth]{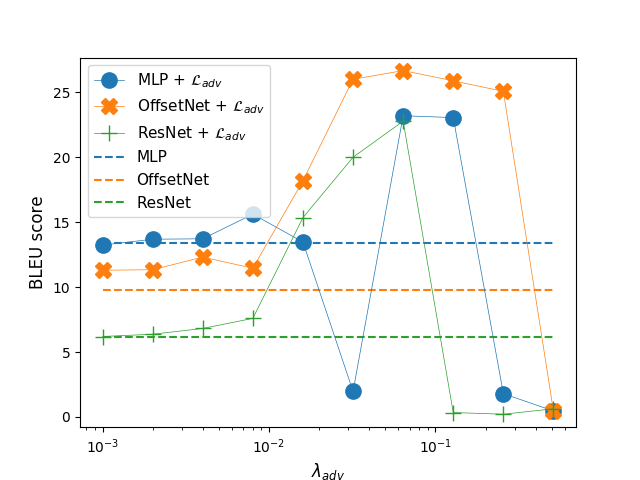}
\vspace{-7mm}
\caption{Performance on WikiLarge in terms of BLEU score on the development set (higher is better) by weight for the adversarial term $\lambdaadv$.  Note that the $x$-axis is on a log scale. %
}
\label{fig:adversarialloss}
\vspace{-3mm}
\end{figure}

From these results, we conclude that, when pretraining a fixed-size-representation autoencoder for plug and play text generation, learning text transformations entirely in continuous space may be easier and more efficient than using conventional sequence-to-sequence models. 

\subsubsection{Sensitivity Analysis} \label{sec:exp:wiki:analyis}

We next explore two of the novel aspects of our model, the adversarial loss and the use of OffsetNet in the mapping function $\Phi$.  We vary the tradeoff parameter $\lambdaadv$ and consider variants of our approach using an MLP, ResNet, and OffsetNet at each value.  All other hyperparameters are kept fixed to default values reported in Appendix~\ref{app:experimental-details}.

\paragraph{Results.}Figure~\ref{fig:adversarialloss} plots the BLEU scores, with $\lambdaadv = 0$ as horizontal dashed lines.
Each model's BLEU score benefits, in some $\lambdaadv$ range, from the use of the adversarial loss.  Gains are also seen for SARI (see Appendix~\ref{app:additional-results}).
OffsetNet is also consistently better than ResNet and, when using the adversarial loss, the MLP.  From this we conclude that OffsetNet's approach of starting close to the input's embedding (and hence on/near the manifold), facilitates (adversarial) training compared to the MLP and ResNet, which, at the beginning of training, map to an arbitrary point in the embedding space due to the randomly initialized projection at the last layer.

\subsection{Sentiment Transfer} \label{sec:style_transfer}

We next evaluate our model on an unsupervised style transfer task.  
For this task, the training data is given pairs of input sentences and sentiment attributes $(\boldsymbol{x}_i, a_i)$, where $\boldsymbol{x}_i$ denotes the input sentence in English and $a_i$ denotes its target sentiment, a binary value. 
For training, we use the Yelp dataset preprocessed following \citet{shen2017style}.
At inference time, we follow common evaluation practices in this task~\cite{hu2017toward, shen2017style,subramanian2018multiple} and evaluate the model on its ability to ``flip'' the sentiment (measured as the accuracy of a DistilBERT classifier trained on the Yelp training set, achieving 97.8\% on held-out data;
\citealp {sanh2019distilbert}),\footnote{Due to budget constraints, we evaluate only on transforming sentiment from negative to positive.} and ``self-BLEU,'' which computes the BLEU score between input and output to measure content preservation.
There is typically a tradeoff between these two goals, so it is useful to visualize performance as a curve (accuracy at different self-BLEU values).
We conduct three experiments.  First, we compare to two other unsupervised sentiment transfer models that can be considered ``plug and play''
(Section~\ref{sec:exp:yelp:plugandplay}).  Second, we conduct controlled experiments with variants of our model to establish the effect of pretraining (Section~\ref{sec:pretraining}). Third, we confirm the effectiveness of OffsetNet and the adversarial loss term for the sentiment transfer (Appendix~\ref{sec:model_analysis}).

\subsubsection{Comparison to Plug and Play Methods}\label{sec:exp:yelp:plugandplay}
To the best of our knowledge, there are only two other autoencoder-based methods that can be used in a plug and play fashion, i.e., training the autoencoder and sentiment transfer tasks in succession. These are the method of \citet{shen2019latent}, which is based on addition and subtraction of mean vectors of the respective attribute corpora (see Section~\ref{sec:problem}), and FGIM (see Section~\ref{sec:inference}); both of them are inference-time methods and do not learn a function (like $\embtransformation$ in Emb2Emb).
\begin{figure}[h]
\centering
\includegraphics[width=0.94\columnwidth]{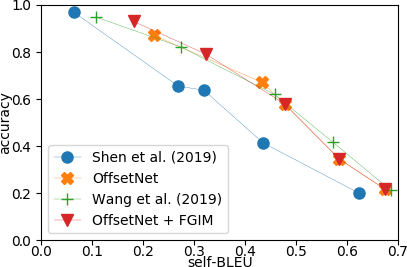}
\vspace{-2mm}
\caption{Comparison of plug and play methods for unsupervised style transfer on the Yelp sentiment transfer task's test set. Up and right is better%
}
\vspace{-2mm}
\label{fig:sentiment-transfer}
\vspace{-2mm}
\end{figure}
 Even though these methods are not specifically introduced with pretraining plug and play in mind, we can consider them in this way as alternatives to our model. %
Note that \citet{wang2019controllable} achieve state-of-the-art on unsupervised sentiment transfer using FGIM, but applied to the latent space of a powerful transformer autoencoder.
Since we want to conduct controlled experiments to find the best plug and play method, we integrated FGIM into our framework rather than directly comparing to their results. We treat the method by \citet{shen2019latent} analogously.

For our learning-based model, we tune $\lambdaadv$ on the development set from Yelp. %
After finding the best $\lambdaadv$, we inspect the behavior of the models at different levels of transfer by varying $\lambdaclf$ ($\{0.1, 0.5, 0.9, 0.95, 0.99\}$), giving a tradeoff curve (more details in Appendix \ref{sec:model_analysis}).
Analogously, %
we vary the multiplier for \citet{shen2019latent} and the thresholds $t$ for \citet{wang2019controllable} to obtain different tradeoffs between accuracy and self-BLEU.
We also report the computational overhead each method incurs in addition to encoding and decoding.

\paragraph{Results.}  Figure~\ref{fig:sentiment-transfer} plots the tradeoff curves for the existing 
models, and ours with and without FGIM at inference time. 
We report accuracy and computational overhead in Table~\ref{tab:pnp}, %
for the most accurate points.
Note that our model clearly outperforms that of \citet{shen2019latent}, confirming that learning the offset vectors in the autoencoder manifold %
is indeed beneficial.\footnote{In Appendix~\ref{app:qualitative-analysis}, we analyze the differences between these models' outputs qualitatively.}\begin{table}[htp]
    \centering
  \def\arraystretch{1.1}\tabcolsep=3.5pt
    \begin{tabular}{l|r|r|r}
        \toprule
        \textbf{Model} & \textbf{Acc.} & \textbf{s-BLEU} & \textbf{+Time} \\
        \toprule
        \hline
        Shen et al. &  96.8 & 6.5 & 0.5$\times$ \\ 
        FGIM & 94.9 & 10.8 & 70.0$\times$ \\ \hline
        Emb2Emb + FGIM & 93.1 & 18.1 & 2820.0$\times$ \\ 
        Emb2Emb & 87.1 & 22.1  & 1.0$\times$ \\ \bottomrule
    \end{tabular}\vspace{-1mm}
    \caption{Self-BLEU (``s-BLEU'') on the Yelp sentiment transfer test set for the configurations in Figure~\ref{fig:sentiment-transfer} %
    with highest transfer accuracy (``Acc.'').  %
    ``+Time'' reports the inference-time slowdown factor due to each model's additional computation (relative to our method).%
    }\vspace{-3mm}
    \label{tab:pnp}
\end{table}  
Our model's performance is close to that of \citet{wang2019controllable}, even without FGIM at inference time. Consequently, our model has a much lower computational overhead. With FGIM, our model shows an advantage at the high-accuracy end of the curve (top), increasing content preservation by 68\% while reaching 98\% of FGIM's transfer accuracy, though this is computationally expensive.
This confirms that our training framework, while being very flexible (see Section~\ref{sec:summary}), is a strong alternative not only in the supervised, but also in the unsupervised case.

\subsubsection{Pretraining} \label{sec:pretraining}

We have argued for a plug and play use of autoencoders because it allows generation tasks to benefit from independent research on autoencoders and potentially large datasets for pretraining.  Here we measure the benefit of pretraining directly by simulating low-resource scenarios with limited style 
supervision.  We consider three pretraining scenarios:
\begin{itemize}
    \item \textbf{Upper bound}: We pretrain on all of the texts and labels; this serves as an upper bound for low-resource scenarios.
    \item \textbf{Plug and play}: A conventional plug and play scenario, where all of texts are available for pretraining, but only 10\% of them are labeled (chosen at random) for use in training $\Phi$.
    \item \textbf{Non plug and play}: A matched scenario with no pretraining (``non plug and play''), with only the reduced (10\%) labeled data.
\end{itemize}

\begin{figure}[h]
\includegraphics[width=0.95\columnwidth]{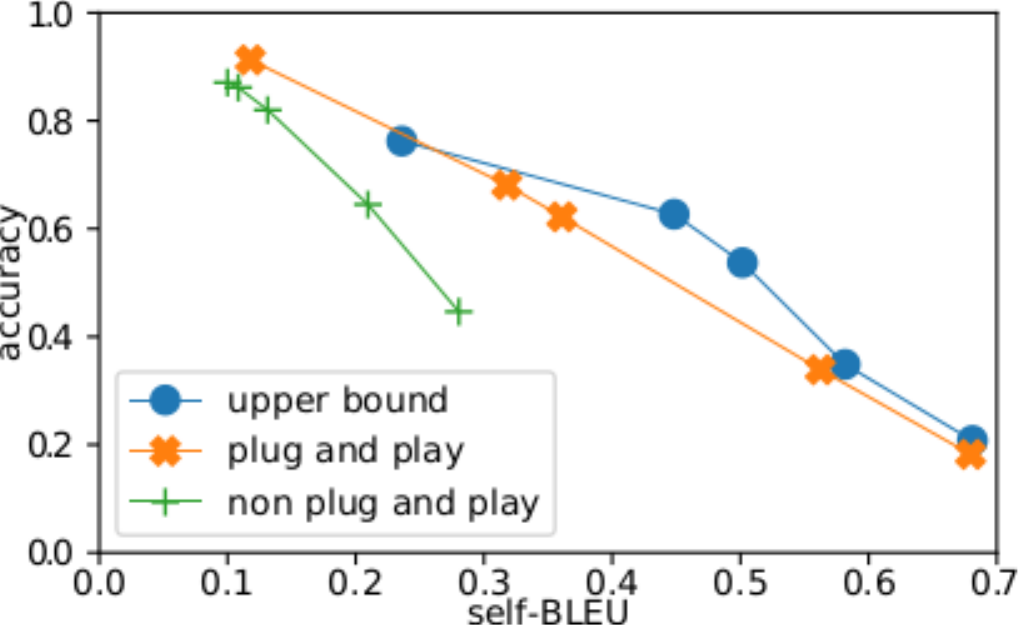}
\vspace{-2mm}
\caption{Sentiment transfer results for different model scenarios. Up and right is better.}
\label{fig:sentiment-transfer-size}
\vspace{-4mm}
\end{figure}

\paragraph{Results.}
Figure~\ref{fig:sentiment-transfer-size} shows the tradeoff curves in the same style as the last experiment.  The benefit of pretraining in the low-resource setting is very clear, with the gap compared to the plug and play approach widening at lower transfer accuracy levels.  The plug and play model's curve comes close to the ``upper bound'' (which uses ten times as much labeled data), highlighting the potential for pretraining an autoencoder for plug and play use in text generation tasks with relatively little labeled data.

\section{Related Work}\label{sec:problem}

\paragraph{Text Style Transfer}

The most common approach to text style transfer is to learn a disentangled shared latent space that is agnostic to the style of the input. Style transfer is then achieved by training the decoder conditioned on the desired style attribute,  ~\cite{hu2017toward, shen2017style, fu2018style, zhao2018adversarially, subramanian2018multiple, li2018delete, logeswaran2018content, yang2018unsupervised, li2019domain}, which hinders their employment in a plug and play fashion.
Most methods either rely on adversarial objectives~\cite{shen2017style, hu2017toward, fu2018style}, retrieval~\cite{li2018delete}, or backtranslation~\cite{subramanian2018multiple, logeswaran2018content} to make the latent codes independent of the style attribute.
Notable exceptions are Transformer-based~\cite{dai2019style, sudhakar-etal-2019-transforming}, use reinforcement learning for backtranslating through the discrete space~\cite{liu2019transformer}, build pseudo-parallel corpora~\cite{kruengkrai2019learning, jin2019imat}, or modify the latent-variable at inference time by following the gradient of a style classifier~\cite{wang2019controllable,liu2019revision}.
Similar to our motivation, %
 \citet{li2019domain} aim at improving in-domain performance by incorporating out-of-domain data into training. However, because their model again conditions on the target data, they have to train the autoencoder jointly with the target corpus, defeating the purpose of large-scale pretraining.

In contrast to previous methods, Emb2Emb can be combined with any pretrained autoencoder even if it was not trained with target attributes in mind.
It is therefore very close in spirit to plug and play language models %
by \citet{dathathri2019plug} who showed how to use pretrained language models for controlled generation without any attribute conditioning (hence, the name). It is also similar to pretrain-and-plugin variational autoencoders~\cite{duan2019pre}, who learn small adapters with few parameters for a pretrained VAE to generate latent codes that decode into text with a specific attribute. 
However, these models cannot be conditioned on input text, and are thus not applicable to style transfer.

 \paragraph{Textual Autoencoders}
Autoencoders are a very active field of research, %
leading to constant progress through denoising~\cite{vincent2010stacked}, variational~\cite{kingma2013autoencoding, Higgins2017betaVAELB, dai2019diagnosing}, adversarial~\cite{makhzani2015adversarial, zhao2018adversarially}, and, more recently, regularized~\cite{Ghosh2020From} autoencoders, to name a few.
Ever since \citet{bowman2015generating} adopted variational autoencoders for sentences by employing a recurrent sequence-to-sequence model, %
 improving both the architecture~\cite{severyn2017hybrid, prato2019towards, liu2019transformer, gagnon2019salsa} and the training objective~\cite{zhao2018adversarially, shen2019latent} have received considerable attention. The goal is typically to improve both the reconstruction and generation performance~\cite{cifka2018eval}.

Our framework is completely agnostic to the type of autoencoder that is used, as long as it is trained to reconstruct the input. Hence, our framework directly benefits from any kind of modelling advancement in autoencoder research.

\section{Conclusion}

In this paper, we present \emph{Emb2Emb}, a framework that reduces conditional text generation tasks to learning in the embedding space of a pretrained autoencoder. We propose an adversarial method and a neural architecture that are crucial for our method's success by making learning stay on the manifold of the autoencoder. Since our framework can be used with any pretrained autoencoder, it will benefit from large-scale pretraining in future research.

\section*{Acknowledgments}
The authors thank Maarten Sap, Elizabeth Clark, 
and the anonymous reviewers for their helpful feedback. Florian Mai was supported by the Swiss National Science Foundation under the project
LAOS, grant number ``FNS-30216''. Nikolaos Pappas was supported by the Swiss National Science Foundation under the project UNISON, grant number ``P400P2\_183911''.  This work was also partially supported by US NSF grant 1562364.

\bibliography{references}
\bibliographystyle{acl_natbib}

\clearpage
\appendix

\section{Experimental Details}\label{app:experimental-details}
We first describe the experimental details that are common to the experiments on both datasets. Dataset-specific choices are listed in their respective subsections.

\subsection{Preprocessing and Tokenization}
We do not apply any further preprocessing to the datasets that we obtain. We use BPE for tokenization, and restrict the vocabulary to 30,000. We truncate all inputs to 100 tokens at maximum.

\subsection{Experimental Setup}
\paragraph{Computing Infrastructure.}
For all of our experiments, we relied on a computation cluster with a variety of different GPUs with at minimum 12GB GPU memory and 50GB RAM.
For the text simplification experiments where we measure training speed, we ran all experiments on the same machine (with a GeForce GTX 1080 Ti) in succession to ensure a fair comparison.

\paragraph{Implementation.}
We used Python 3.7 with PyTorch 1.4 for all our experiments. %
Our open-source implementation is available at \url{https://github.com/florianmai/emb2emb}.

\paragraph{Adversarial Training.}
We employ a 2-layer MLP with 300 hidden units and ReLU activation as discriminator, and train it using Adam with a learning rate of 0.00001 (the remaining parameters are left at their PyTorch defaults). We train it in alternating fashion with the generator $\embtransformation$, in batches of size 64.

\subsection{Neural Architectures}
\paragraph{Encoder}
For encoding, we employ a one-layer bidirectional LSTM as implemented in PyTorch. To obtain the fixed-size bottleneck, we average the last hidden state of both directions. The input size (and token embedding size) is 300.

\paragraph{Decoder}
For decoding, we initialize the hidden state of a one-layer LSTM decoder as implemented in PyTorch with the fixed size embedding. During training, we apply teacher forcing with a probability of 0.5. The input size is 300. We use greedy decoding at inference time.

\paragraph{Transformation $\embtransformation$.}
We train all neural network architectures with one layer. The hidden size is set to the same as the input size, which in turn is determined by the size of the autoencoder bottleneck. Hence, the MLP and OffsetNet have the same number of parameters. Due to its extra weight matrix at the output-layer, the ResNet has 50\% more parameters than the other models. All networks use the SELU activation function.
All training runs with our model were performed with the Adam optimizer.

\subsection{Text Simplification}\label{sec:sim_details}

\subsubsection{Dataset Details}
We evaluate on the WikiLarge dataset by \citet{zhang2017sentence}, which consists of sentence pairs extracted from Wikipedia, where the input is in English and the output is in simple English.
It contains of 296,402 training pairs, 2,000 development pairs, and 359 pairs for testing. The 2,359 development and test pairs each come with 8 human-written reference sentences to compute the BLEU and SARI overlap with.
The dataset can be downloaded from \url{https://github.com/XingxingZhang/dress}.

\subsubsection{Experimental Details}
\paragraph{Training our model.}
We use a fixed learning rate of 0.0001 to train our model for 10 epochs. We evaluate the validation set performance in terms of BLEU after every epoch and save the iteration with the best validation loss performance.

\paragraph{Training S2S models.}
For all S2S models we compare against in Section~\ref{sec:exp:wiki:end2end}, we select the best performing run on the validation set among the learning rates $\{0.001, 0.0005, 0.0001, 0.00005, 0.00001, 0.000005\}$, and also assess the validation set performance after each of the 20 epochs. Training is performed with the Adam optimizer.

\paragraph{Encoder hyperparameters}
We use a 1-layer bidirectional LSTM with a memory size of 1024 and an input size of 300.

\paragraph{Number of Parameters}
All models share the same encoder and decoder architecture, consisting of 34,281,600 parameters in total. The mappings MLP, OffsetNet, and ResNet have 2,097,132, 2,097,132, and 3,145,708 parameters, respectively.
We report total numbers for the models used in the experimental details below.

\paragraph{Evaluation metrics.}
For computing BLEU, we use the Python NLTK 3.5 library.\footnote{\url{https://www.nltk.org/api/nltk.translate.html##module-nltk.translate.bleu_score}} For computing SARI score, we use the implementation provided by ~\cite{xu2016optimizing} at \url{https://github.com/cocoxu/simplification/blob/master/SARI.py}.

\subsubsection{SARI Score by $\lambdaadv$}\label{app:additional-results}

\paragraph{Experimental details.}
We measure the performance of our model on the development set of WikiLarge in terms of SARI score. These results are for the same training run for which we reported the BLEU score, hence, the stopping criterion for early stopping was BLEU, and we report the results for all 10 exponentially increasing values of $\lambdaadv$.
The best value when using BLEU score as stopping criterion is $\lambdaadv = 0.032$.

\begin{figure}[h]
\includegraphics[width=1.07\columnwidth]{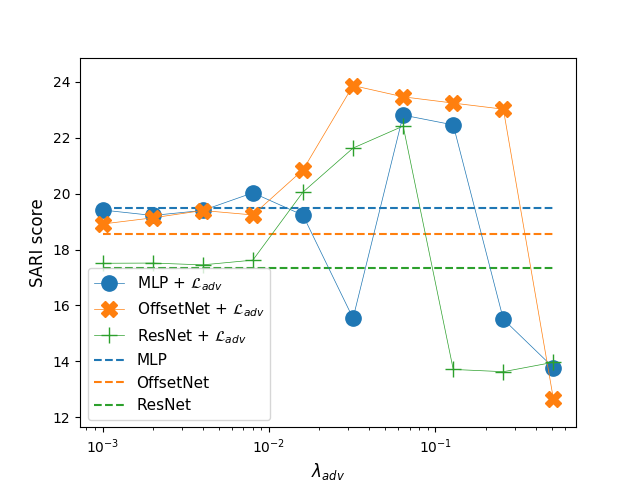}
\caption{Performance on WikiLarge in terms of SARI score (higher is better) by weight for the adversarial term $\lambdaadv$.}
\label{fig:adversarialloss-sari}
\end{figure}

\paragraph{Results.}
The results in Figure~\ref{fig:adversarialloss-sari} show the same pattern as for the BLEU score, although with a smaller relative gain of 23\% when using the adversarial term.

\subsubsection{Development Set Results for Comparison to S2S Models}
In Table~\ref{tab:end2end-developmentset}, we report the development set performances corresponding to the experiments reported in Section~\ref{sec:exp:wiki:end2end}. For each model, we also specify the best learning rate, if applicable, and the number of parameters in the model
\begin{table}[htp]
    \centering
  \def\arraystretch{1.1}\tabcolsep=4pt    
    \begin{tabular}{l|r|r|r|c}
        \toprule
        \textbf{Model} & \textbf{BLEU} & \textbf{SARI} & \textbf{LR} & $\mathbf{|\Theta|}$ \\
        \toprule
        \hline
        \etoescratch & 3.2 & 14.3 & 0.0001 & 34.3m \\
        \etoept & 5.9 & 15.1 & 0.0005 & 34.3m \\
        \etoemlp & 8.6 & 16.0 & 0.0001 & 36.4m \\
        \etoefreeze & 17.4 & 20.1 & 0.00005 & 36.4m \\
        \hline
        Ours & \textbf{26.7} & \textbf{23.5} & - & 36.4m \\ \bottomrule
    \end{tabular}
    \caption{Text simplification performance of model variants of seq2seq training on the development set. $|\Theta|$ denotes the number of parameters for each model.
    }
    \label{tab:end2end-developmentset}
\end{table}
\subsection{Sentiment Transfer}

\subsubsection{Dataset Details}
We evaluate on the Yelp dataset as preprocessed by \cite{shen2017style}, which consists of sentences with positive or negative sentiment extracted from restaurant reviews.
The training set consists of 176,787 negative and 267,314 positive examples. The development set has 25,278 negative and 38,205 positive examples, and the test set has 50,278 negative and 76,392 positive examples. 
The dataset can be downloaded from \url{https://github.com/shentianxiao/language-style-transfer/tree/master/data/yelp}.

\paragraph{Training our models.}
We use a fixed learning rate of 0.00005 to train our model for 10 epochs (for the ablations) or 20 epochs (for the final model). We evaluate the validation set performance in terms of self-BLEU plus transfer accuracy after every epoch and save the iteration with the best validation loss performance.

For all models involving training the mapping $\embtransformation$ (including the ablation below), we perform a search of $\lambdaadv$ among the values $\{0.008, 0.016, 0.032, 0.064 0.128\}$.
We select them based on the following metric: $\sum\limits_{i = 1}^5 (BLEU(\lambdaadv, \lambdaclf^i) + accuracy(\lambdaadv, \lambdaclf^i)$, where $\lambdaclf^i$ corresponds to the $i$-th value of $\lambdaclf$ that we have used to obtain the BLEU-accuracy tradeoff curve. By $BLEU(\lambdaadv, \lambdaclf^i)$ and $accuracy(\lambdaadv, \lambdaclf^i)$, respectively, we mean the score resulting from training with the given parameters.

\paragraph{Encoder hyperparameters}
We use a 1-layer bidirectional LSTM with a memory size of 512.

\paragraph{Number of Parameters}
All models again share the same encoder and decoder architecture, consisting of 22,995,072 parameters in total. The mappings MLP, OffsetNet, and ResNet have 524,288, 524,288, and 786,432 parameters, respectively. Hence, the total number of parameters for our models is 23.5m, whereas the variants we report as Shen et al. and FGIM have 23m parameters.

\paragraph{Sentiment classifier on autoencoder manifold.}
For binary classification, we train a 1-layer MLP with a hidden size of 512 with Adam using a learning rate of 0.0001. For regularization, we use dropout with $p = 0.5$ at the hidden and input layer, and also add isotropic Gaussian noise with a standard deviation of 0.5 to the input features.

\paragraph{BERT classifier.}
The DistilBERT classifier is trained using the HuggingFace transformers library.\footnote{Specifically, we use the run\_glue.py script in from \url{https://github.com/huggingface/transformers} and only replace the SST-2 dataset with the Yelp dataset. We used the commit ``11c3257a18c4b5e1a3c1746eefd96f180358397b'' for training our model.} We train it for 30 epochs with a batch size of 64 and a learning rate of 0.00002 for Adam, with a linear warm-up period over the first 3000 update steps. We evaluate the validation set performance every 5000 steps and save the best model.

\subsubsection{Implementation of Wang et al. Baseline}
We reimplemented the Fast Gradient Iterative Modification method by \cite{wang2019controllable} to either i) follow the gradient of the sentiment classifier from the input, or ii) from the output of $\embtransformation$, follow the gradient of the complete loss function of training $\embtransformation$.

Following the implementation by \cite{wang2019controllable}, in all runs, we repeat the computation for weights $\omega \in \{1, 10, 100, 1000\}$ and stop at the first weight that leads to the classification probability exceeding a threshold $t$. For each weight, we make 30 gradient steps at maximum.

The \citet{wang2019controllable} baseline is generated from choosing $t = \{0.5, 0.9, 0.99, 0.999, 0.9999\}$, i.e., we choose lower thresholds to stop the gradient descent from changing the input too much towards the target attribute, leading to lower transfer accuracy performances.

When we apply FGIM to the output of $\embtransformation$ in our model (with the more sophisticated loss function, where we set $\lambdaclf = 0.5$), we apply the same thresholds.

\subsubsection{Development Set Result for Comparison of Plug and Play}

\begin{figure}[h]
\centering
\includegraphics[width=0.94\columnwidth]{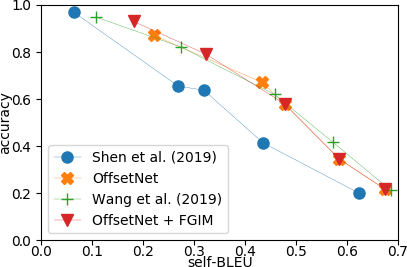}
\caption{Comparison of plug and play methods for unsupervised style transfer on the Yelp sentiment transfer task's development set. Up and right is better%
}
\label{fig:sentiment-transfer-dev}
\end{figure}

In Figure~\ref{fig:sentiment-transfer-dev}, we report the development set result corresponding to the test set results of the experiments presented in Section~\ref{sec:exp:yelp:plugandplay}. These results are shown for $\lambdaadv = 0.008$, which performed the best in terms of the development score metric introduced in the training details.

\subsubsection{Model Analysis} \label{sec:model_analysis}

\paragraph{Experimental Setup}
We investigate the effect of OffsetNet and the adversarial training term on our unsupervised style transfer model by measuring the self-BLEU score with the input sentence and the accuracy of a separately trained BERT classifier (achieving 97.8\% classification accuracy) on the Yelp development set. We again report the best performance among 6 exponentially increasing $\lambdaadv$ values for each model. To inspect the behavior of the models at varying levels of transfer, we trained and plotted one model each for $\lambdaclf \in \{0.1, 0.5, 0.9, 0.95, 0.99\}$.

\begin{figure}[h]
\includegraphics[width=0.95\columnwidth]{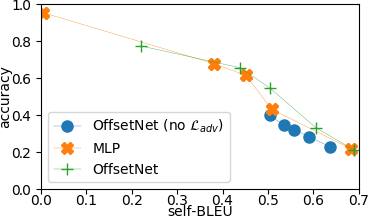}
\caption{Ablation of our model components on the Yelp sentiment transfer tasks. Up and right is better.}
\label{fig:sentiment-transfer-ablation}
\end{figure}

\paragraph{Results.}
The results in Figure~\ref{fig:sentiment-transfer-ablation} show that OffsetNet reaches better transfer accuracy than the MLP at comparable self-BLEU scores. The performance drops significantly if the adversarial term is not used.
This confirms the importance of our design decisions.

\section{Qualitative Analysis}\label{app:qualitative-analysis}

We provide several example outputs of our method in comparison to the outputs of the baseline by \citet{shen2019latent} in Tables~\ref{tab:ex0}, \ref{tab:ex1}, \ref{tab:ex2}, and \ref{tab:ex3}. %
Moreover, we show how the output evolves as the multiplier and $\lambdaclf$ (i.e., the level of transfer accuracy) increases. 

In our qualitative analysis we generally observe that both models generate similar outputs when the inputs are short and can be transferred by only changing or deleting single words (e.g., Table~\ref{tab:ex0}). We observe that grammaticality degrades in both methods for higher transfer levels. 
However, our method is more often able to preserve the content of the input as the transfer accuracy increases: At a multiplier of 3.0, the method by \citet{shen2019latent} outputs rather general positive statements that are mostly disconnected from the input, whereas our method is able to stay on the topic of the input statement. This observation matches the quantitative results from Section~\ref{sec:exp:yelp:plugandplay}, where our method attains substantially higher self-BLEU scores at comparable levels of transfer accuracy.

However, it is clear that both models mostly rely on exchanging single words in order to change the sentiment classification. In the example from Table~\ref{tab:ex1}, our model changes the input ``the cash register area was empty and no one was watching the store front .'' to the rather unnatural sentence ``the cash area was great and was wonderful with watching the front desk .'' instead of the more natural, but lexically distant reference sentence ``the store front was well attended ''. We think that this is best explained by the fact that we use a denoising autoencoder with a simple noise function (deleting random words) for these experiments, which encourages sentences within a small edit-distance to be close to each other in the embedding space~\cite{shen2019latent}. Denoising autoencoders with a more sophisticated noise functions focused on semantics could possibly mitigate this, but is out of scope for this study.

\begin{table*}[ht]
    \centering
    \begin{tabulary}{\textwidth}{|C
    |L|L|}
    \hline
    multiplier / $\lambdaclf$ & Shen et al. (2019) & Ours \\
    \hline
    1.5 / 0.5 & i will be back . & i will be back . \\
    \hline
    2.0 / 0.9 & i will be back back &i will definitely be back .  \\
     \hline
    2.5 / 0.95 &  i will definitely be back . & i will definitely be back  \\
     \hline
    3.0 / 0.99 & i love this place ! & i will be back ! \\
    \hline
    \end{tabulary}
    \caption{\textbf{Input}: i will never be back .}
    \label{tab:ex0}
\end{table*}

\begin{table*}[ht]
    \centering
    \begin{tabulary}{\textwidth}{|C
    |L|L|}
    \hline
    multiplier / $\lambdaclf$ & Shen et al. (2019) & Ours \\
    \hline
    1.5 / 0.5 & the cash area was great and the the best staff & the cash area was great and was wonderful one watching the front desk . \\
    \hline
    2.0 / 0.9 & the cash register area was empty and no one was watching the store front . & the cash area was great and was wonderful with watching the front desk . \\
     \hline
    2.5 / 0.95 & the cash bar area was great and no one was the friendly staff . & the cash area was great and was wonderful with watching the front desk . \\
     \hline
    3.0 / 0.99 & the great noda area and great and wonderful staff . & the cash area was great and her and the staff is awesome ! \\
    \hline
    \end{tabulary}
    \caption{\textbf{Input}: the cash register area was empty and no one was watching the store front . \textbf{Reference}: the store front was well attended}
    \label{tab:ex1}
\end{table*}

\begin{table*}[ht]
    \centering
    \begin{tabulary}{\textwidth}{|C|L|L|}
    \hline
    multiplier / $\lambdaclf$ & Shen et al. (2019) & Ours \\
    \hline
     1.5 / 0.5 & we sit down and we got some really slow and lazy service .  & we sit down and we got some really slow and lazy service . \\
    \hline
    2.0 / 0.9 & we sit down and we got really awesome and speedy service .  & we sit down and we got some really slow and lazy service .  \\
     \hline
    2.5 / 0.95 & we sit down and we we grab the casual and and service .  & we sit down and we got some really great and and awesome service .  \\
     \hline
    3.0 / 0.99 & we sit great and and some really great and awesome atmosphere .  & we sit down and we got some really comfortable and and service . \\
    \hline
    \end{tabulary}
    \caption{\textbf{Input}: the cash register area was empty and no one was watching the store front . \textbf{Reference}: the service was quick and responsive}
    \label{tab:ex2}
\end{table*}

\begin{table*}[]
    \centering
    \begin{tabulary}{\textwidth}{|C|L|L|}
    \hline
    multiplier / $\lambdaclf$ & Shen et al. (2019) & Ours \\
    \hline
      1.5 / 0.5 & definitely disappointed that i 'm not my birthday ! & definitely disappointed that i could not use my birthday gift ! \\
    \hline
     2.0 / 0.9 & definitely disappointed that i have a great ! & definitely not disappointed that i could use my birthday gift ! \\
     \hline
     2.5 / 0.95 & definitely super disappointed and i 'll definitely have a great gift ! & definitely disappointed that i could use my birthday gift ! \\
     \hline
     3.0 / 0.99 & definitely delicious and i love the ! & definitely disappointed that i could use my birthday gift !\\
    \hline
    \end{tabulary}
    \caption{\textbf{Input}: definitely disappointed that i could not use my birthday gift ! \textbf{Reference}: definitely not disappointed that i could use my birthday gift !}
    \label{tab:ex3}
\end{table*}

\end{document}